\documentclass[11pt]{article}
\usepackage[preprint]{acl}
\usepackage{fontspec}
\newfontfamily\cjkfont{FandolSong-Regular.otf}[BoldFont=FandolSong-Bold.otf]
\newcommand{\cn}[1]{{\cjkfont\XeTeXlinebreaklocale "zh"\XeTeXlinebreakskip=0pt plus 1pt #1}}
\usepackage{amsmath,amssymb,booktabs,tabularx,array,graphicx}
\definecolor{teal}{HTML}{167D8D}
\definecolor{amber}{HTML}{D69535}
\definecolor{rose}{HTML}{B45F6A}
\hypersetup{colorlinks=true,linkcolor=teal,citecolor=teal,urlcolor=teal,pdftitle={Long-Lived Characters, Local Inference: Incremental Memory Maintenance for Game NPCs},pdfauthor={Zimu Xu}}

\newcommand{\code}[1]{\texttt{#1}}
\newcommand{\takeaway}[1]{\par\smallskip\noindent\textbf{Takeaway.} #1\par\smallskip}
\newcommand{\quotebox}[2]{\begin{quote}\small\textbf{#1}\par\smallskip #2\end{quote}}
\newcolumntype{Y}{>{\raggedright\arraybackslash}X}
\title{Long-Lived Characters, Local Inference:\\Incremental Memory Maintenance for Game NPCs}
\author{Zimu Xu\\University of Bern\\\href{mailto:zimu.xu@unibe.ch}{\texttt{zimu.xu@unibe.ch}}}
\date{}

\begin{document}
\raggedbottom
\maketitle
\begin{abstract}
A game character should not have to reread its entire life before every conversation. For locally deployed language-model characters, however, revising a few memories can invalidate a long reusable prefix. The resulting preparation cost competes with both foreground dialogue and the maintenance of other characters. This matters especially when dialogue feeds game-defined actions and value judgments: a fluent but incorrect account of who owns an item, or whether a transfer has already happened, can corrupt the input to otherwise deterministic rules. We study incremental memory maintenance for long-lived game NPCs in a quantized Qwen hybrid recurrent--attention model. Our runtime removes superseded attention KV entries, computes replacement records at the true sequence tail, and preserves the continuing recurrent state and unchanged KV. Existing local experiments combine multi-update dialogue replays, fixed-input placement ablations, and attention diagnostics. Independent block composition weakens query-conditioned memory selection without a uniform chunk-initial attention collapse. True-tail updates preserve important current-state and historical bindings across eight scripted maintenance rounds; a placement case recovers the full-refill quantity in three reconstructions, while slot-preserving alternatives repeat a double-subtraction error. Attention-distribution proximity alone does not explain these semantic differences. The results motivate treating a character's inference state as a maintained, history-dependent resource, rather than only a disposable encoding of its latest memory text.
\end{abstract}

\section{Introduction}
\begin{figure*}[t]
\centering
\includegraphics[width=\textwidth]{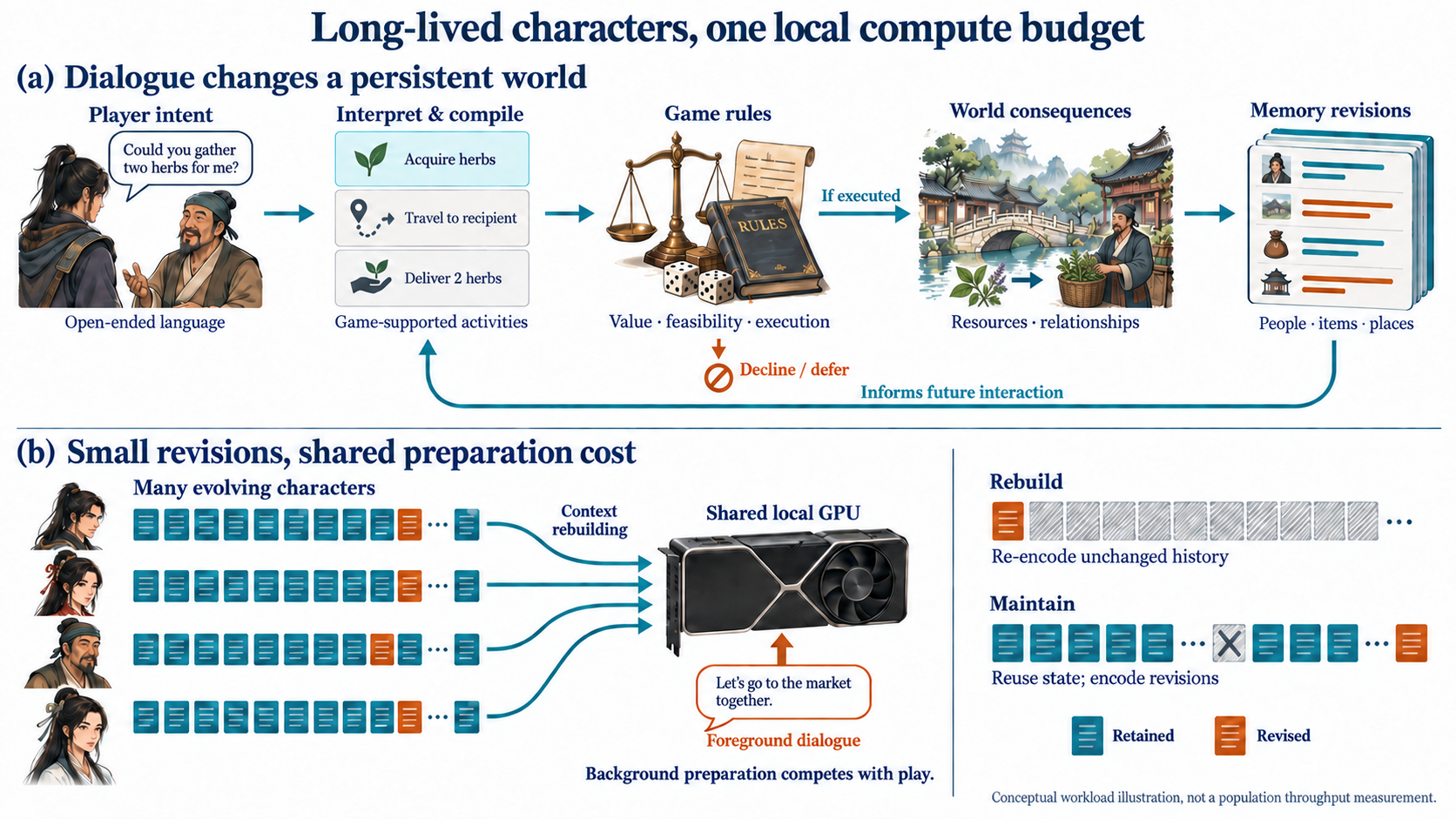}
\caption{\textbf{Rule-governed play makes memory maintenance consequential.} Open-ended dialogue expresses intentions; compilation connects them to game-supported activities, while game rules govern authorization and consequences. Experiences revise character records. On a shared local device, repeatedly rebuilding long contexts competes with foreground dialogue. The target is to encode revisions while reusing surviving state. This conceptual illustration motivates the workload; it is not a population-throughput measurement or a claim that the complete envisioned game has been implemented.}
\label{fig:motivation}
\end{figure*}
An NPC that can produce a convincing sentence is not yet an NPC on which a game can depend. Players make plans because they expect resources, relationships, and consequences to retain meaning beyond the current exchange. A promised gift is not an executed transfer. A remembered act of kindness is not evidence that a former ally remains safe. A herb already given away must not be subtracted from the current inventory a second time. These distinctions become mechanically important when open-ended dialogue connects to the systems that govern a game.

Our design premise is that \emph{gameplay should be governed by controllable rules, not improvised anew by model generation}. Language models interpret flexible requests and character-specific circumstances. An action compiler translates relevant intentions into game-supported activities, while local value and execution rules determine their mechanical consequences. Subjective preferences may differ between characters, and uncertainty may be deliberately introduced through mechanisms such as dice. Neither requires abandoning a coherent economic scale or letting an eloquent sentence create an item. Section~\ref{sec:design} explains this premise without assuming that the complete envisioned game has already been implemented or evaluated.

The resulting character needs more than a static persona. It must revise beliefs about people, objects, and places while retaining relevant experience. Memory can be compressed or merged; a seller's claim can be contradicted by a later test; ownership can change without changing an entity's identity. Generative-agent research has established the usefulness of experience, reflection, and retrieval for believable behavior \citep{park2023generative}. Conversational memory systems likewise manage what to retain and bring into active context \citep{zhong2024memorybank,packer2023memgpt}. Here we focus on a complementary deployment problem: \emph{how can an evolving character context remain computationally available on a player's machine?}

In one measured local configuration, preparing the full dialogue pipeline from cold took 57.5 seconds. This is multi-prefix preparation, not the generation time of one reply. An individual wait may be hidden by prefetching. If dozens of characters need refreshed contexts, however, asynchronous scheduling moves the work without removing it. On a shared local accelerator, background rebuilding competes with the conversation the player is actually having. A memory system that is inexpensive to edit as text can therefore remain expensive to use as a model state. Figure~\ref{fig:motivation} connects this preparation budget to the persistent consequences that make dialogue useful to a game.

Prefix caching does not, by itself, solve this problem. Systems such as SGLang reuse shared prefixes across generation calls \citep{zheng2024sglang}, but updating an early record invalidates the exact reusable prefix from that point onward. Modular reuse goes beyond this restriction \citep{gim2024promptcache,yao2025cacheblend}; the challenge becomes which contextual computation to preserve or repair. Keeping every version as appended text avoids replacement but grows the visible history and preserves conflicting versions. Independently computing blocks avoids some rebuilding, yet breaks the contextual computation that originally related them. Hybrid models add another complication: token-indexed attention KV coexist with recurrent and short-convolution states that cannot be deleted one record at a time.

We investigate an alternative: retain the continuing recurrent state, remove the direct KV of superseded records, and compute replacements at the true tail. The logical memory is editable even though its computational history keeps moving forward. This is intentionally not an exact reconstruction of the latest text. We ask whether it can preserve the \emph{semantic continuity} needed by a game character while avoiding recomputation of unchanged records.

Our contributions are threefold:
\begin{itemize}
\item We formulate a persistent-memory workload for \emph{fully local game characters}, connecting memory revision to rule-governed dialogue, resource accounting, and action binding.
\item We implement incremental hybrid-state maintenance that separates the lifetime of explicit KV records from the continuing recurrent state, without changing model weights.
\item We present a case-study and diagnostic evaluation of independent composition, true-tail maintenance, and position-preserving alternatives. The results expose both selective-memory addressing failures and state--event binding errors that aggregate attention similarity does not reliably predict.
\end{itemize}

The evidence in this study is deliberately organized by experimental cohort, rather than presented as a single benchmark leaderboard. It includes one synthetic character under multiple histories and controlled module replays; multi-character population evaluation remains future work.

\section{Design Goal: A Playable Dialogue Interface}
\label{sec:design}
\subsection{Game-defined value before stochasticity}
Open-ended language is useful because the player can express intentions beyond a fixed dialogue menu. Its value to a game depends on what those intentions can do. Trading, bargaining, deception, and relationship-building become strategic when the player can learn which conditions matter and trust that investments have persistent meaning.

Our intended value system connects costs and benefits through game-defined exchange relationships: resources, opportunities, and character-specific preferences must be interpretable on a coherent scale. This does not imply identical prices or perfectly rational NPCs. Trust, need, or attachment may change willingness to pay. The distinction is between a preference with maintained reasons and an arbitrary change of economic scale caused by a new generation. Fairness means that the rules governing such differences are sufficiently stable for the player to form a strategy.

Likewise, controllability does not imply a deterministic world. A dice check has a trigger, a probability model, and defined consequences. It adds uncertainty \emph{within} a mechanism. Uncontrolled model variability should not silently replace that mechanism. The language model may portray hesitation, pride, or dishonesty; the game must still distinguish what was said, what was intended, what was authorized, and what actually happened.

\subsection{The harness connects meaning to mechanics}
At a high level, the character harness separates three responsibilities. \textbf{Interpretation} resolves the active matter and its context. \textbf{Compilation} describes the character's potential game-supported activities and, where necessary, the entities and quantities involved. \textbf{Adjudication} applies the game's value, feasibility, and execution rules. Dialogue communicates the resulting response or plan. The language used in a promise is not itself an execution receipt.

This separation supports a specific product direction: the player can influence a character through flexible language, but cannot simply narrate away resource costs. Proposed mechanics such as disguise-based conversations or costly memory intervention illustrate why the interface is more than a chat window: open language enables unscripted tactics, while world rules determine their boundaries. These mechanics motivate the architecture; their entertainment value and full implementation are not claims of the present experiments.

\subsection{Memory continuity is a mechanical dependency}
Even a deterministic rule receives the wrong input if the character confuses the owner of an item, an intended transfer with a completed one, or a former ally with a current threat. We therefore study memory as a dependency of game-defined decision-making, not merely as a source of colorful anecdotes.

Consider three statements: ``I collected three herbs,'' ``I already gave one away,'' and ``I currently hold two.'' They describe one consistent history, not three independent quantities to combine freely. A small binding error can change a compiled transaction while leaving the prose entirely plausible. This is the kind of error the evaluation prioritizes. Multiple natural replies and reasonable guesses remain acceptable; factual relationships that determine mechanical inputs do not become interchangeable.

\section{Problem: Mutable Memory on a Shared Local Device}
\label{sec:problem}
\subsection{Records, revisions, and character sessions}
A character's memory store contains experience records and entity-cognition records. Experiences include a time description, summary, and retained details. Cognition records identify people, items, and places and describe the character's current understanding of them. The same entity ID can occur in several memories; revising its description does not create a new person or item.

Let $\mathcal M_t$ be the live record set after maintenance step $t$. An update supplies records $U_t$ and identifies superseded records $D_t$:
\begin{equation}
\mathcal M_{t+1}=(\mathcal M_t\setminus D_t)\cup U_t.
\end{equation}
This supports additions, replacement, and many-to-one compression. A revision may remove descriptive detail while preserving event participants and current consequences. The experiments use scripted updates so that the intended facts and lost details are auditable. They do not evaluate an autonomous memory editor.

Three orders need not coincide: when an event happened, when its description was last revised, and where its current KV representation lies. An old experience can receive a recent revision; a recently relevant experience can remain in old, unchanged KV. Successful maintenance must not confuse these orders.

\subsection{The character-readiness budget}
We distinguish three costs. \emph{Cold preparation} builds reusable contexts and warms the modules. \emph{Maintenance} revises these contexts after new experience. \emph{Hot interaction} processes the current suffix and produces a visible reply. Reducing the third does not remove the first two.

Fully local deployment makes their competition concrete. If $N$ characters each require a standalone rebuild costing roughly $c$ device-seconds, $Nc$ is a useful first-order accounting of serial work, not a prediction of parallel wall-clock latency. Scheduling, shared prefixes, and batching can change realized costs. They cannot make repeated computation free. A world with many changing characters needs to reduce work as well as hide it. We treat this as deployment motivation; no multi-NPC throughput measurement is reported here.

\subsection{Semantic continuity, not output identity}
For a query $q_t$, a maintained state should support outputs consistent with the relevant live facts and legitimate uncertainty in $\mathcal M_t$. We do not require the same wording, mood, or speculative explanation as a fresh refill. We distinguish:
\begin{itemize}
\item \textbf{State and event fidelity:} quantities, ownership, participants, temporal relations, and completed versus proposed actions.
\item \textbf{Contextual usability:} whether recent revisions and relevant unchanged memories can inform dialogue and compilation.
\item \textbf{Character expression:} coherent, intelligible dialogue, allowing more than one reasonable interpretation or strategy.
\end{itemize}
Fresh refill is a computational reference, not an infallible narrative oracle. It can make mistakes too.

This emphasis connects to long-term conversational evaluation: LoCoMo studies temporally structured experience, while LongMemEval explicitly tests knowledge updates and temporal reasoning \citep{maharana2024locomo,wu2025longmemeval}. RULER further distinguishes basic retrieval from tracing and aggregation in long contexts \citep{hsieh2024ruler}. These works motivate evaluating relationships, not just finding a record. Our scripted replays are not runs of these benchmarks.

\section{Incremental Hybrid-State Maintenance}
\label{sec:method}
\begin{figure*}[t]
\centering
\includegraphics[width=\textwidth]{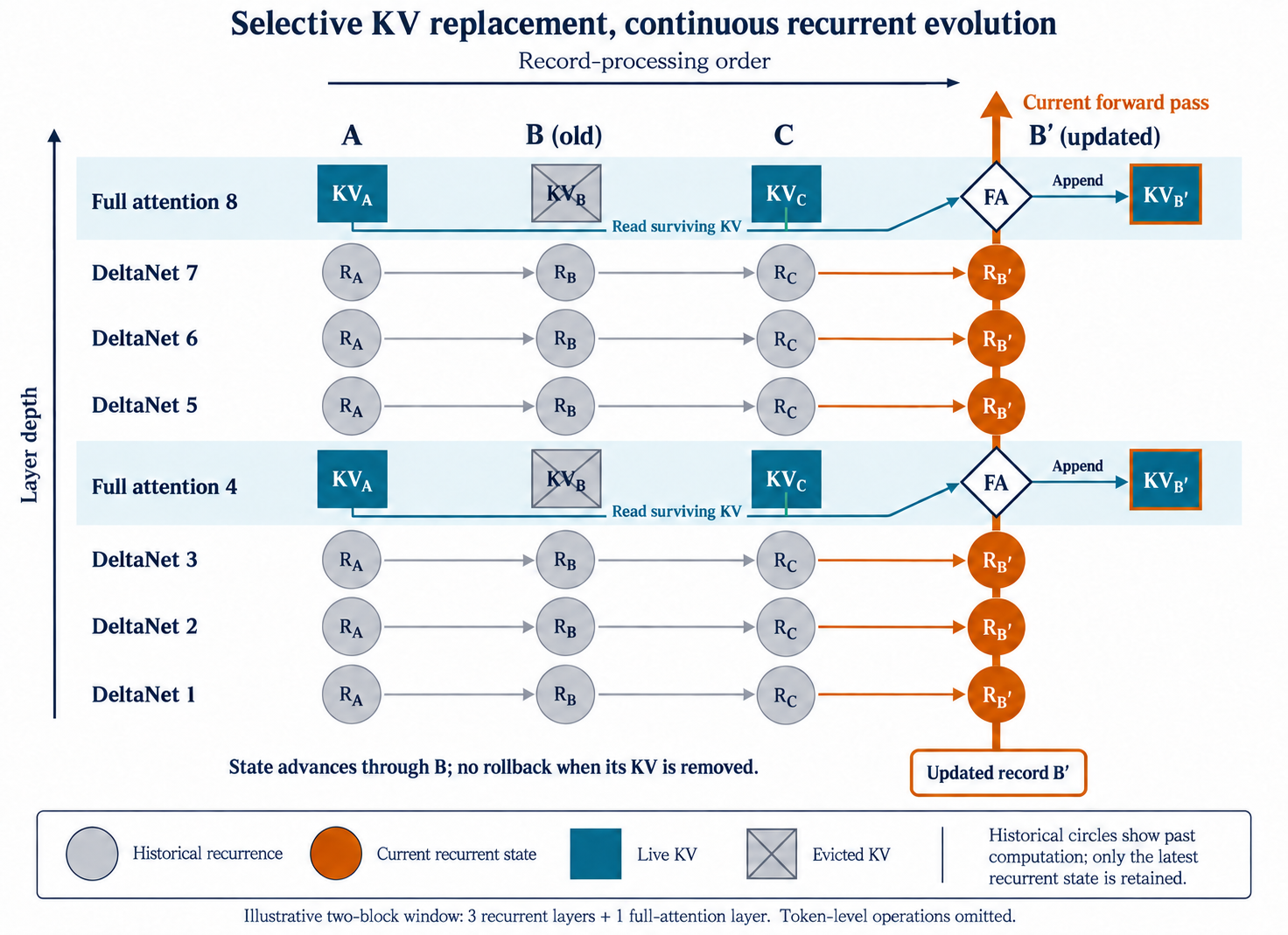}
\caption{\textbf{One update, two state lifecycles.} Columns show record-processing order, not event time; rows show model depth. Historical circles depict past recurrent computation, while only each layer's latest state is retained. The orange vertical path carries the replacement's current hidden activations through DeltaNet updates and full-attention blocks (diamonds), not through KV storage (squares). Superseded B KV are removed, surviving A/C KV are read, and new B$'$ KV are appended at the true tail. Recurrence and convolution state continue without rollback. Two hybrid blocks are shown; each contains three recurrent layers and one full-attention layer. Deletion does not erase indirect historical influence. This is schematic AI-generated artwork; experimental attention charts use recorded data.}
\label{fig:concept}
\end{figure*}
\subsection{Two state structures, two lifecycles}
The evaluated model interleaves Gated DeltaNet recurrent layers with full-attention layers \citep{yang2025gated,qwen2026model}. Figure~\ref{fig:concept} shows their different state lifecycles during a record replacement. Denote the live attention cache by $A_t$, the recurrent and convolution states by $R_t$, and the next sequence-position ID by $p_t$. A character context is therefore not just a token list but a state $(A_t,R_t,p_t)$.

Full refill computes a new state from the latest rendered text. Our method instead advances an existing state. After invalidating the direct KV of superseded records, the update is conceptually
\begin{align}
\bar A_t &= A_t\setminus A_t[D_t],\\
(A_{t+1},R_{t+1},p_{t+1})
&=F_\theta(U_t;\bar A_t,R_t,p_t),
\end{align}
where $F_\theta$ applies the ordinary forward computation through the complete hybrid model, with record-span bookkeeping and temporary maintenance delimiters whose direct KV are subsequently removed. Replacement records are evaluated at their actual new tail positions. They can read surviving KV, while the recurrent state continues from its previous value. The model weights are frozen.

The recurrent state may retain influence from removed text. So may surviving deeper-layer KV that were computed in its presence. Removing a record eliminates its direct future attention access; it does not uncompute its history. We make this distinction explicit rather than describe the operation as exact forgetting.

\subsection{Record-addressed runtime operations}
Each live record has a stable semantic identifier and a runtime mapping to its token spans. Cache-version identifiers do not enter the model's text. The runtime verifies deletion ranges against live spans, removes replaced entries, and retains unrelated KV without replaying their tokens. Brief maintenance delimiters may be processed to contextualize a revision and then removed from retained KV; their indirect computational effects remain.

The true-tail route never rotates updated keys back into the old record's position. Rotary position embedding makes query--key interactions position-dependent \citep{su2021roformer}; moving a stored key is therefore not merely moving an entry in a table. Deleted positions become holes, not dummy KV tokens. Live-cache occupancy and the monotonically advancing position cursor are different quantities. Position IDs continue growing even when the amount of live KV remains bounded. Managing very long positional trajectories is a separate unresolved limit, not a benefit claimed from the deletion operation.

Different tasks can have different cached prefixes. The prototype maintains separate recurrent histories for the main character context and the three action-analysis stages. Task-specific continuations branch from their appropriate maintained roots. We do not merge unrelated task states into one recurrent matrix, nor claim that every early pipeline module uses this maintenance path.

\subsection{What computation is avoided}
Unchanged records are not recomputed during an update. The remaining work includes encoding replacement text, its attention against surviving context, state and cache management, and any necessary suffix rebuilding. Thus the saving is not a claim that update time depends only on the number of changed tokens: attending to a larger live context still costs work.

The practical distinction is between repeatedly encoding all records and repeatedly accessing their retained representations. This is especially relevant when a small revision invalidates a long exact-match prefix. The current implementation demonstrates that unchanged-token recomputation can be eliminated in tested maintenance traces; a complete population-level latency advantage requires additional measurement.

\section{Experimental Design}
\label{sec:experiments}
\subsection{A local game-character prototype}
The recorded runs use a Qwen3.6-27B PRISM PRO DQ derivative in GGUF form, served through a modified local \code{llama.cpp} runtime. The model family uses 48 recurrent layers interleaved with 16 full-attention layers; the latter have 24 query heads in the instrumented build. We cite the base-family model card for architectural context, not as the identity of the derivative weights \citep{qwen2026model,llamacpp}. No weight training is performed in these experiments.

Historical runtime logs identify the checkpoint as Q3\_K--Medium, with 27.32 billion parameters. The local device has an NVIDIA GeForce RTX 5090 Laptop GPU with 24,462 MiB VRAM, an Intel Core Ultra 9 285HX, and approximately 64 GiB system RAM. The recorded server configuration allocates 49,152 live KV token slots, uses GPU layer offload, and sets batch and microbatch sizes to 2,048 and 512. These are the measured prototype's settings, not minimum deployment requirements.

The synthetic character, Lin Qingyao, interacts with the player character Yin Xu in a Chinese-fantasy setting. Earlier experiments use 34 memory groups with 123 retained details: the memory region is 13,250 tokens, with 1,699 tokens of common head context. A later scenario interleaves memory and cognition records and tests changes to item effects, inventory, destinations, and a formerly trusted relative. We translate example excerpts into English and retain decisive Chinese originals in Appendix~\ref{app:examples}.

Early independent-composition experiments use MTP=0. The eight-update mixed-record replay and the later placement ablations use MTP=1 for applicable upstream modules; production dialogue follows its target-only streaming path. Comparisons are made within a recorded configuration, not across MTP settings as if only the cache method had changed.

\subsection{Cohorts and comparison discipline}
\begin{table*}[t]
\centering\small
\begin{tabularx}{\textwidth}{@{}p{0.16\textwidth}p{0.25\textwidth}Y@{}}
\toprule
Evidence cohort & Interventions & What is held fixed / what it supports \\
\midrule
Independent composition & Full refill vs independently computed memory groups; two dialogue turns & First-turn request is shared. Second turns consume their own preceding replies. Separate attention probes freeze the suffix and continuation. \\
Five-update attention & True-tail maintenance vs refill of the final 38-group text & Same final text, record order, suffix, and fixed continuation; supports a paired reading-pattern comparison, not a free-generation accuracy rate. \\
Eight-update replay & 25 mixed records maintained into 24; nine full dialogue turns & One maintained trajectory with real downstream outputs. No same-final-text full-refill trajectory was rerun. \\
10K placement probes & Dense refill; fresh 10K slots; maintained original slots & Three frozen module requests: dialogue, herb compilation, and compilation gating. Same final text, ordering, and suffix within this three-way comparison. \\
Tail / relocation ablation & True-tail 10K updates; tail computation then key rotation back & Frozen herb-compilation task. Tail updates change final record order and positions. Three reconstructions each; not three independent test cases. \\
Long-horizon prompt probes & Original vs understanding-first task on a restored maintained state & Three frozen module inputs, one response per arm: two affect probes and one dialogue probe. Only the final task changes; records, history, and incoming module outputs are fixed. \\
\bottomrule
\end{tabularx}
\caption{\textbf{Existing evidence is complementary, not a pooled leaderboard.} Experimental identifiers and source reports are listed in Appendix~\ref{app:provenance}. ``10K'' denotes a logical position interval, not ten thousand occupied KV cells per record.}
\label{tab:cohorts}
\end{table*}

Table~\ref{tab:cohorts} separates end-to-end trajectories from fixed-input diagnostics. This matters because an early change can alter a reply, which then changes every downstream context. Such a trajectory demonstrates usability but cannot localize a later error to cache placement alone. Frozen-module replays address the narrower question by keeping history and upstream outputs fixed.

We compare six constructions. \textbf{Dense refill} reads the current text sequentially. \textbf{Independent composition} computes each memory group from the same public head, at its final position, then assembles its attention KV; recurrent and convolution states resume from the head before the suffix is read. This is a naive independent-block experiment, not a reimplementation of CacheBlend or EPIC. \textbf{Fresh gapped prefill} sequentially reads the final text with 10K sequence-position intervals between records. \textbf{Slot-preserving maintenance} advances recurrence at the causal work tail but evaluates replacement Q/K using RoPE positions offset to the original slot, then relabels cached positions. \textbf{True-tail maintenance} gives updates their actual new tail positions. \textbf{Tail-and-relocate} computes updates using tail-position RoPE, then rotates stored keys back to old positions, leaving values and recurrent state otherwise unchanged.

\subsection{Attention measurement}
We instrument actual query, key, and mask tensors at full-attention layers while retaining Flash Attention in the runtime. Attention probabilities are reconstructed with a float32 reference softmax. The main diagnostics average the last eight query tokens at each probe, then aggregate heads and layers as specified. Logical holes are excluded from the live-token denominator. The 48 recurrent layers have no corresponding softmax attention map, although they affect the measured queries and keys.

We distinguish absolute attention mass from attention normalized within the memory records. A heatmap can look similar after normalization while the model allocates a different total share to memory. For group probabilities $a_i$, $\exp(-\sum_i a_i\log a_i)$ is reported as the \emph{effective group count}; it describes dispersion, not the number of facts understood. Distribution distances are diagnostic summaries, not causal explanations of generated errors. The debate on attention-based explanation motivates separating descriptive traces from causal attribution \citep{jain2019attention,wiegreffe2019attention}.

\section{Results}
\label{sec:results}
\subsection{Independent composition weakens selective addressing}
Independent composition does not make all memory unreadable. Both initial dialogue turns complete, and the character still accesses the discovery location, injuries, and later care. It also develops a plausible hypothesis linking different events. Nevertheless, selective addressing changes markedly.

\begin{figure*}[t]
\centering
\includegraphics[width=\textwidth]{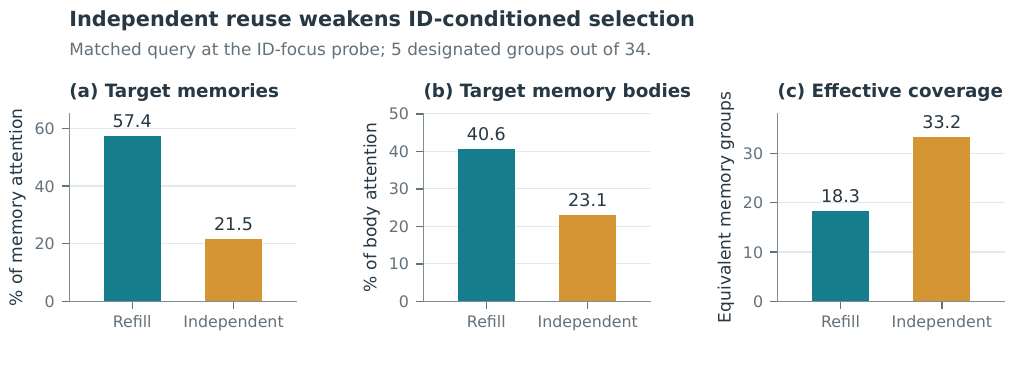}
\caption{\textbf{More dispersed memory attention can mean weaker task-directed selection.} Paired probes after the same focus-ID instruction; the five requested groups comprise about 12.16\% of memory tokens. Effective group count is an entropy-based dispersion measure. These measurements are not generated-answer accuracy scores.}
\label{fig:selectivity}
\end{figure*}

At the explicit focus-ID probe, five requested memory groups receive 57.4\% of memory attention under full refill but 21.5\% under independent composition. Their share of memory-body attention falls from 40.6\% to 23.1\%, while the effective number of attended groups rises from 18.3 to 33.2 out of 34 (Figure~\ref{fig:selectivity}). The distribution becomes nearly uniform across groups at this probe.

At the reply-start probe, total memory attention increases by approximately 9.75 percentage points, of which 9.06 points are attributable to IDs and structural tokens. This is not equivalent to a comparable increase in factual content reading. At the fixed factual continuation, both states can again concentrate on relevant memory bodies: the five groups receive 67.2\% and 57.4\% of body attention, respectively. The problem is weakened selection in particular contexts, not universal memory loss.

We do not observe a uniform chunk-initial collapse. The first eight tokens of every group account for 5.86\% versus 6.03\% of memory attention at reply start, and 0.53\% versus 0.19\% at the factual continuation. At the ID probe their share does increase, from 5.58\% to 7.63\%, but ID lookup gives this boundary region a legitimate semantic role. These observations differ from the pervasive chunk-start pattern analyzed by EPIC. The constructions also differ: EPIC's analysis independently encodes chunks from position zero, whereas our blocks share a head and are computed at their final positions. We therefore report a different observed signature, not a refutation of EPIC's setting \citep{hu2025epic}.

A corresponding end-to-end failure appears in the first response plan. The memory explicitly places the discovery at dawn today. The independently composed plan instead says to explain that Yin Xu was found last night. The full-refill response uses this morning. The final independent dialogue contains an ellipsis that admits a conversational reading; the unambiguous evidence is the planner's explicit event--time binding (Appendix~\ref{app:examples}). The attention probes use a frozen reference suffix, not the instant that erroneous plan was generated, so this is a co-occurring failure example rather than a demonstrated causal chain.

\takeaway{Independent composition can retain broadly coherent dialogue while weakening query-conditioned memory selection and allowing a local factual binding error. Chunk-head attention alone is not a sufficient diagnostic.}

\subsection{Tail maintenance retains relevant old and revised records}
\begin{figure*}[t]
\centering
\includegraphics[width=\textwidth]{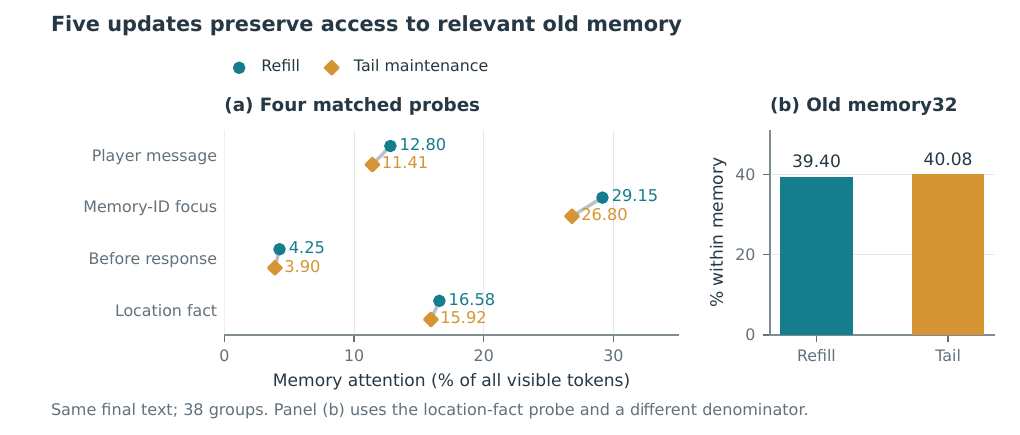}
\caption{\textbf{An old KV record remains available after repeated tail updates.} Five-update maintenance and full refill of the same final text are compared at four frozen probes. The old-record detail concerns memory32 at the location-relation probe. Similarity here is a measured reading pattern, not proof of identical hidden state or generated wording.}
\label{fig:continuity}
\end{figure*}

In the paired five-update experiment, maintenance leaves 38 live memory groups. Both arms see the same final text and task suffix. At the location-relation probe, total memory attention is 16.58\% for refill and 15.92\% for maintenance; memory-body attention is 13.04\% and 12.56\%. An unchanged older record, memory32, receives 39.40\% and 40.08\% of all memory attention. Recent appended records do not prevent the query from recovering this older, relevant material (Figure~\ref{fig:continuity}).

The mixed-record replay provides complementary behavioral evidence. Eight updates revise one or two cognition entries at a time and merge related memories lossily. For example, a purchased charm changes from a seller's claimed protection to an observed harmful effect and is finally retained as evidence. A trusted relative becomes a dangerous attacker while the older rescue experience remains true. Herb quantities and the recipient of a cord change. Other experiences are never updated.

All nine dialogue turns reach a final response. The four turns requiring selective third-stage entity binding produce concrete bindings without unresolved entries: the false protective charm, two remaining herbs, the player's cord, and the attacking relative. The character also distinguishes the old kindness from the new danger and retrieves who arrived first, owned a book, and repaired it in an unchanged older memory. A representative response is: ``The warmth in the past was real; the danger now is real too.''

This is not a flawless trajectory. One compilation stage drops a daytime restriction that the preceding stage had retained. The last dialogue invents an unsupported cultivation rank for the attacker, although the target entity binding is correct. There is also ambiguity in a rescuer reference and repetition after repeated value refusals. These are reported because successful retrieval and parameter binding do not certify every downstream sentence. Without a paired refill trajectory for this cohort, they are not attributed specifically to maintenance.

\takeaway{Eight scripted, lossy maintenance rounds preserve several demanding current-state and historical bindings through a real dialogue pipeline. The evidence supports practical feasibility, not universal error-free behavior.}

\subsection{Update placement changes state--event binding}
The herb example isolates a particularly important failure: treating a current quantity as a quantity \emph{before} an already-completed transfer. The maintained history records three collected herbs, one already given away, and two remaining. The player's request is to give the remainder.

\begin{table}[t]
\centering\small
\begin{tabularx}{\columnwidth}{@{}Ycc@{}}
\toprule
Computation path & Quantity & Rebuilds \\
\midrule
Dense refill & 2 & 1 \\
Fresh prefill + 10K slots & 2 & 1 \\
Maintain original 10K slots & 1 & repeated$^{\dagger}$ \\
Maintain true tail + 10K & 2 & 3/3 \\
Tail compute, rotate back & 1 & 3/3 \\
\bottomrule
\end{tabularx}
\caption{\textbf{Current-state versus historical-event binding.} All outputs concern the same frozen herb-compilation task. $^{\dagger}$Original-slot failures recur in the initial and subsequent control rebuilds; no pooled success rate is inferred. Three reconstructions are repeatability checks, not independent scenarios. Tail placement also changes record ordering.}
\label{tab:herb}
\end{table}

Dense refill and fresh gapped prefill both compile two herbs. Maintaining the old slots instead compiles one. Three independent reconstructions with true-tail updates all compile two, with identical outputs; three tail-and-relocate reconstructions all compile one. The latter explicitly states: ``According to memory, I have two warm-pulse herbs in total; I gave one to Granny He yesterday, so one remains.'' It has applied a past deduction to a quantity that already incorporates it.

The final live cache contains 11,894 tokens in the placement comparison. Its next-position metadata is 261,955 for the original-slot layout and 479,547 for the true-tail layout. The holes consume no placeholder KV. Thus this is not a comparison between 12K and 480K occupied token caches.

One stale cognition phrase says the character does not currently hold the herb, although its updated quantity and the revised memory say two. This imperfection is shared by the matched fixtures. The example therefore tests robust integration of current quantities and historical events in the presence of a stale description, rather than pristine arithmetic. The output difference remains real, but the fixture should not be represented as completely unambiguous.

Rotating keys back is insufficient in this example. It changes the stored positional component of keys but does not recompute values, deeper contextual representations, or the recurrent history under the destination arrangement. True-tail ordering, positions, recency, and computational history are coupled in this experiment. The evidence supports the operational choice of true-tail updates; it does not identify one of those factors as the sole cause.

\subsection{Attention proximity does not certify semantic fidelity}
The separate three-way attention experiment compares dense refill, fresh gapped prefill, and original-slot maintenance, with identical final text, order, history, and upstream outputs. It does \emph{not} contain a true-tail attention capture.

\begin{figure*}[t]
\centering
\includegraphics[width=\textwidth]{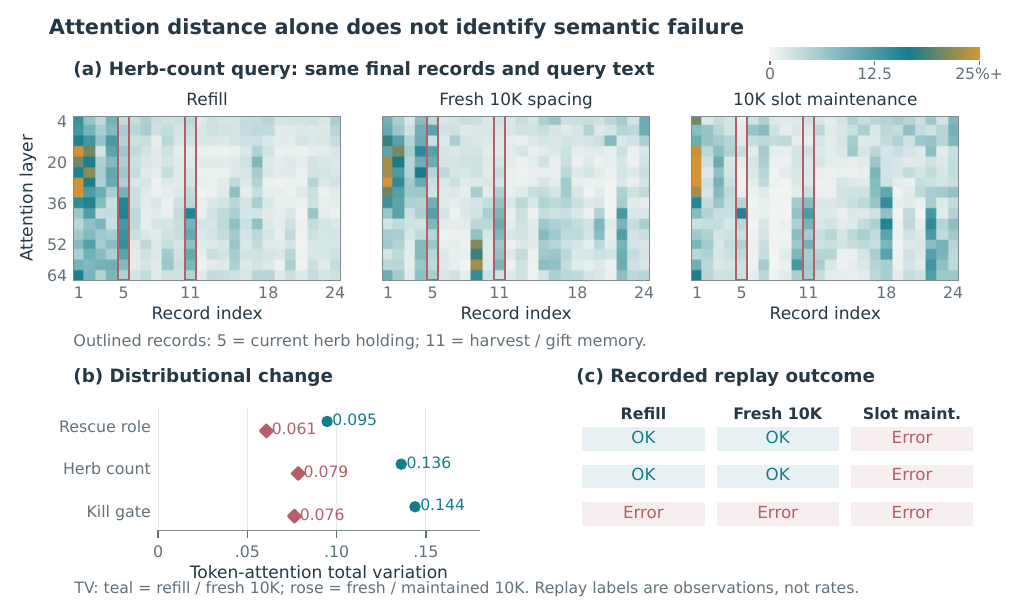}
\caption{\textbf{Distributional proximity is not a semantic correctness certificate.} Three-way fixed-text diagnostics compare dense refill, fresh 10K-gapped prefill, and original-slot maintenance. Attention is captured before output, averaging the final eight query tokens; behavior markers refer to the corresponding recorded module replays, not new generations during tracing. The true-tail ablation in Table~\ref{tab:herb} is a different experiment and is not plotted as an attention arm.}
\label{fig:placement}
\end{figure*}

Across dialogue, herb compilation, and compilation gating, token-distribution total variation distances from dense to fresh-gapped prefill are 0.095, 0.136, and 0.144. Distances from fresh-gapped prefill to maintained-gapped state are smaller: 0.061, 0.079, and 0.076. Yet the first transition preserves the correct rescued-person reference and herb quantity, while the second is accompanied by mistakes in both (Figure~\ref{fig:placement}). A larger measured distribution shift can therefore coexist with a correct answer, and a smaller one with a discrete binding error.

Nor does a single ``less attention to evidence'' explanation fit every error. The herb-related cognition and memory receive less absolute attention under maintenance. For the rescued-person dialogue, however, attention to the relevant record is higher than under fresh gapped prefill, despite the wrong participant being spoken. The compilation-gating request fails in all three paths; it is not evidence of a maintenance-specific regression.

These are case-level diagnostics, not a statistical test of all attention-distance metrics. They nevertheless caution against using a visually similar heatmap, or one aggregate distance, as a substitute for auditing quantities and event roles.

\subsection{Prompt-guided recovery without rebuilding memory}
\label{sec:prompt-recovery}
A long-horizon experiment supplies a complementary observation: a factual binding error can be corrected while retaining the maintained memory state. The trajectory spans eight simulated years and 121 supplied experiences, with model-generated memory and cognition revisions. We examine frozen downstream requests from this trajectory rather than reporting its dialogue turns as independent accuracy trials.

The decisive case asks who picked up Lin Qingyao's scattered notebook pages. The retained cognition explicitly says that Yin Xu collected and cleaned them. Nevertheless, the original affect task reverses the actor: ``I remember that I picked up the scattered pages.'' It also imports a red cord from a different episode into its explanation of the character's feelings.

Both replay arms resume the same 11,301-token maintained root, with identical preceding messages and incoming state. The original-task control reproduces the actor reversal. A generic understanding-first task instead asks the model to establish how the interaction developed from the available memories, cognition, and dialogue, then derive its feelings from that understanding rather than rearrange events to explain those feelings. It supplies no case-specific answer or entity checklist, adds no call or reasoning field, and preserves the output schema and MTP setting. The resulting explanation correctly states: ``That day, he carefully picked them up and brushed them clean for me,'' without importing the cord (Appendix~\ref{app:prompt-recovery}). Only request suffixes are prefilled; the maintained root is neither rebuilt nor repaired.

This is evidence of \emph{prompt-steerable factual access}: the maintained state still supports the correct relation, and task framing changes whether it is used correctly. The result is consistent with directing task-level attention toward event understanding before affective elaboration. It is a behavioral intervention, not a measured attention-map shift. Across the three paired probes, the other affect error did not recur in its control, while the dialogue-location error persisted after its task change; this finding establishes a recoverable case rather than a general repair rate.

\takeaway{A binding error after maintenance need not imply irreversible memory loss. In the same retained state, a general change in task organization can recover the correct event relation without rebuilding the prefix.}

\subsection{Update cost follows the changed records}
For one approximately 11.9K-token activity-compilation prefix, initial construction takes 10.216 seconds. Eight subsequent mixed memory-and-cognition revisions take only 3.481 seconds in total. Each revision supplies 214--362 changed-record tokens and costs 0.338--0.490 seconds, including its temporary maintenance framing. No unchanged record is re-encoded. Thus, all eight revisions together cost less than one initial prefix construction.

\begin{figure*}[t]
\centering
\includegraphics[width=\textwidth]{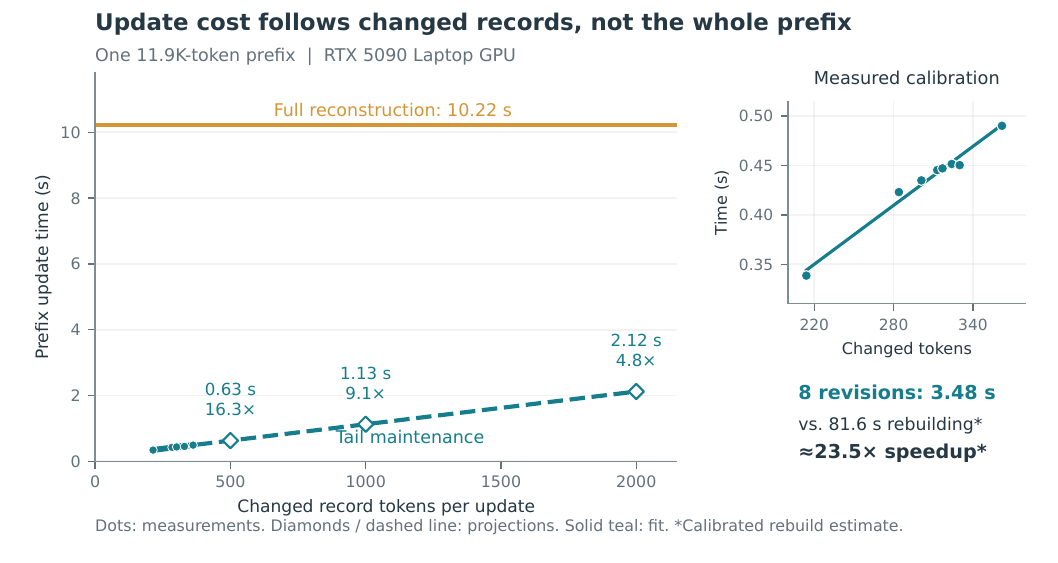}
\caption{\textbf{Same-prefix update cost, calibrated on a consumer laptop.} The reconstruction reference is the same root's measured initial construction: 11,852 tokens in 10.216 seconds. Teal dots are eight sequential measured revisions; the inset shows their calibration range. An affine fit is solid within that range and dashed when extrapolated to larger replacements, assuming similar record structure and a roughly fixed live-prefix size. Maintenance timings include an 11-token wrapper excluded from the horizontal-axis count. For the eight-revision total, reconstruction is estimated separately at each observed live length $N_i$ as $10.216\,N_i/11{,}852$, yielding 81.64 seconds versus 3.48 seconds measured maintenance, or approximately $23.5\times$ calibrated speedup. Revision-text generation, response generation, and the final task suffix are outside both update budgets.}
\label{fig:update-scaling}
\end{figure*}

Figure~\ref{fig:update-scaling} relates changed-record volume to update time. A first-order fit to the eight recorded updates gives
\begin{equation}
\widehat{T}_{\mathrm{maint}}(m)=0.131+0.000994m\quad\text{seconds},
\end{equation}
where $m$ counts tokens in the replacement records, not deleted tokens. Holding the prefix size near 11.9K tokens, the calibrated curve projects approximately 0.63 seconds for 500 changed tokens and 1.13 seconds for 1,000, compared with 10.22 seconds for reconstruction. This is the practical benefit of selective replacement: preparation tracks what changed rather than repeatedly processing the character's unchanged history.

Across the observed eight-revision workload, rebuilding after each revision would cost an estimated 81.64 seconds at the measured same-root construction rate, versus 3.48 seconds of maintenance: approximately $23.5\times$ calibrated speedup, or 95.7\% less update time. The final task suffix adds 0.053 seconds. These savings arise from avoiding unchanged-prefix reconstruction; they do not require faster response decoding.

\paragraph{Character readiness and hot interaction.}
That cohort's complete cold preparation takes 48.15 seconds, including initial construction and other module warm-up. This is a separate pipeline-level budget, not the denominator of the same-prefix comparison. The nine prepared dialogue turns average 23.18 seconds to the first visible character and 25.83 seconds to completion, excluding cold preparation and between-turn background preparation.

In the earlier 34-group comparison, full-refill preparation takes 57.466 seconds and cold independent composition 65.375 seconds. Independent composition is not faster to initialize in that prototype. Persisted independent states later reload across a service restart in 5.783 seconds, but that is a storage-reuse result, not an update result; Windows file caches may remain warm. These distinctions matter for a fully local system: reuse, incremental revision, and cold construction solve different parts of character readiness.

\section{Related Work}
\label{sec:related}
\paragraph{Persistent characters and memory management.}
Generative Agents combines experience storage, retrieval, reflection, and planning to produce believable behavior \citep{park2023generative}. MemoryBank adds conversational-memory maintenance with time- and importance-sensitive forgetting \citep{zhong2024memorybank}; MemGPT orchestrates movement between memory tiers and active context \citep{packer2023memgpt}. These systems address which experience remains available. Our complementary question is how revised resident records become computationally available without rebuilding their entire context. Game-defined resource and action rules make this continuity consequential; this motivation is not a claim that prior interactive systems lack consequences.

\paragraph{Learned internal memory.}
LongMem uses a frozen backbone to encode history and a trained side network to retrieve and read cached memory \citep{wang2023longmem}. It illustrates that internal representations can support long-term access, but through a trained architecture. Here the weights are fixed: the intervention is in the runtime lifecycle of an existing hybrid model's KV and recurrent state, not a learned memory reader.

\paragraph{Serving and shared-prefix reuse.}
PagedAttention addresses KV allocation, fragmentation, and sharing within and across requests \citep{kwon2023pagedattention}. SGLang's RadixAttention reuses KV across structured generation programs \citep{zheng2024sglang}. Efficient allocation and reuse are complementary to record maintenance: they do not by themselves specify how to retain a character's computational history after an earlier fact changes. Our workload adds logical revision to the serving problem rather than replacing memory allocation or scheduling.

\paragraph{Position-independent cache reuse.}
Prompt Cache defines reusable modules and manages their positions through a schema \citep{gim2024promptcache}. CacheBlend selectively recomputes cached representations to recover contextual interactions between independently encoded chunks \citep{yao2025cacheblend}. EPIC's LegoLink concentrates repair near chunk beginnings, motivated by its attention-sink analysis \citep{hu2025epic}. We share the concern that raw independent reuse changes computation, but our principal operation is an in-place \emph{logical revision} implemented by true-tail state evolution. We do not implement their full algorithms as baselines, and our diagnostic independent composition should not be labeled ``CacheBlend.''

\paragraph{Streaming retention and positional sensitivity.}
StreamingLLM retains initial attention-sink tokens alongside a recent window for efficient streaming \citep{xiao2024streamingllm}. Our deletion policy instead follows semantic supersession: an old record may stay live while a newer, superseded one is removed. Long-context studies also show that the position of relevant information can change task performance \citep{liu2024lostmiddle}. This motivates testing placement rather than assuming layout is neutral, but neither result identifies the cause of our hybrid-state binding failures.

\paragraph{Hybrid recurrent--attention caching.}
State-space duality connects recurrent and attention formulations \citep{dao2024ssm}, while Gated DeltaNet combines gating with delta-rule state updates \citep{yang2025gated}. At the caching level, HYPIC composes cached segment transition operators and states and repairs cross-segment effects with seam computation \citep{liu2026hypic}. LinearKV decouples recurrent-state initialization from full-attention cache reuse and studies single cached-state initializers \citep{liu2026linearkv}. These works make recurrent handling an explicit design choice for hybrid PIC. Our study instead retains a character's own continuing recurrent state through a sequence of replacements and lossy memory revisions. The distinction is the maintained history and workload, not a claim that hybrid caching or state decoupling is new.

\section{Discussion and Scope}
\label{sec:discussion}
\paragraph{A memory state is not just a cached document.}
Two states can expose the same final records yet differ in how those records were computed. Exact refill is one reference point, but game usability concerns whether the state supports the right entities, quantities, and relationships. The working hypothesis emerging from these experiments is that preserving a coherent forward update path may be more useful than forcing replacement KV back into a familiar textual layout. The present ablations support this hypothesis operationally, without establishing a universal architectural law.

\paragraph{Deletion is access control, not complete erasure.}
Retaining recurrence is a benefit only insofar as its historical influence remains useful. It can also preserve obsolete beliefs. The method is unsuitable as a guarantee that secrets or personal information have been forgotten. It must be evaluated under adversarial revisions, repeated reversals, and longer histories before stronger forgetting or stability claims are made.

\paragraph{What this study establishes.}
We demonstrate an implemented runtime mechanism, working multi-update traces, and concrete placement-sensitive semantic failures. We do not establish a population accuracy rate, a user-study improvement in enjoyment, an end-to-end economic balance guarantee, or superiority to fully implemented PIC systems. The mixed replay includes downstream errors. Quantization, numerical paths, limited repeated trials, and the selected character setting constrain generalization. Repeated outputs in one case are repeatability evidence, not independent samples.

\paragraph{A local deployment agenda.}
Fully local execution gives the game a bounded, player-owned compute budget rather than an elastic remote serving pool. The relevant goal is not merely a faster individual answer. It is keeping a changing cast of characters ready to participate without repeatedly spending that budget on unchanged lives. We offer a concrete prototype and diagnostic evidence toward this goal; the game-facing demonstration is not publicly released with this preprint.

\section{Practical Constraints and Path Forward}
\label{sec:roadmap}
\subsection{Research on a single consumer laptop}
The implementation and experiments reported here were developed and run on a single consumer laptop with an NVIDIA GeForce RTX 5090 Laptop GPU and approximately 24\,GiB of reported device memory (Section~\ref{sec:experiments}). Within this budget, the prototype runs a quantized 27B-class hybrid model, implements direct KV and recurrent-state interventions, and completes character-dialogue replays after repeated memory revisions. This setting connects the research prototype to its intended deployment on player-owned hardware.

The scale of the present evaluation reflects the research funding and compute currently available. Broader access to GPU resources and support for sustained development would allow a wider range of character histories, longer controlled runs, and additional model configurations. At this stage, the study concentrates on implemented state operations, controlled replays, and inspectable failure cases; larger-scale validation forms part of the next phase.

Expanding this evaluation requires more than generating additional replies. The interventions depend on an instrumented runtime with direct access to attention KV, recurrent state, and sequence positions; they cannot be reproduced simply by issuing more requests to a black-box model API. Each character also needs a coherent revision history and an auditable record of which facts remain true. Reconstructing comparison states, replaying long maintenance trajectories, and checking downstream bindings all consume development and evaluation resources. These constraints explain the current scope; broader reliability remains an empirical question.

\subsection{Near-term research priorities}
The planned next stage has three priorities:
\begin{enumerate}
\item \textbf{Broader character histories.} Evaluate a small set of distinct synthetic characters with shared entities, changing relationships, ownership transfers, and lossy memory merges. Include both revised facts and unchanged memories that become relevant again, rather than testing only the newest record.
\item \textbf{Longer and better-controlled maintenance.} Extend the number of revisions, introduce repeated reversals and contradictory reports, and compare maintained states with full-refill references at selected checkpoints. Where resources permit, vary quantization and model configurations while separating those changes from the cache intervention.
\item \textbf{The shared-device readiness budget.} Measure preparation, update work, live-state storage, and foreground reply latency separately. Then test how maintaining several characters competes with active dialogue on the same local device.
\end{enumerate}

The semantic audit will distinguish explicit factual contradictions from unresolved references, defensible inference, and expression choices. Blinded model-assisted review can help, with human adjudication of disagreements and retained source evidence. Known position, verbosity, and self-preference biases motivate explicit rubrics rather than treating model verdicts as ground truth \citep{zheng2023judge}. Human judgments likewise need an auditable factual basis. This roadmap describes planned work, not completed evaluation.

\subsection{Opportunities for collaboration}
This direction offers opportunities for collaboration across hybrid-model research, local-inference systems, and persistent game characters. Shared compute resources, access to additional hardware, and research funding would help extend the present case study into a more systematic evaluation. Independent replication, runtime instrumentation, and character histories with well-specified factual changes would be particularly valuable contributions.

We welcome discussions with researchers and partners interested in developing this direction together. Collaboration may focus on runtime and evaluation artifacts without requiring public release of the game-facing prototype. The longer-term objective remains persistent, rule-governed characters on player-owned hardware; additional research resources would help establish the conditions under which this becomes reliable and practical.

\section{Conclusion}
Long-lived game characters need continuity in both their remembered world and the computation that makes that world available. We study a training-free maintenance route that replaces explicit attention KV while preserving a continuing recurrent history. Existing experiments show usable historical and current-state bindings after repeated lossy updates, weaker selective addressing under independent composition, and placement-sensitive failures that attention similarity alone does not explain. True-tail maintenance is the strongest operational candidate among the tested update paths, without being equivalent to full refill or proven reliable for arbitrary histories. The broader lesson is that persistent local characters require treating memory maintenance as a first-class inference operation, grounded in the stable rules that make dialogue worth playing.

\section*{AI Assistance}
AI tools, including GPT-6 Astra Ultra, assisted with implementation and experiment scripting, analysis of recorded artifacts, and manuscript drafting and revision. Image-generation tools were used for the schematic motivation and architecture illustrations; empirical figures are generated from recorded numerical data and explicitly identified calibrations. The human author is responsible for the scientific claims, source verification, and final submitted content.

\bibliography{references}

\begin{thebibliography}{24}
\providecommand{\natexlab}[1]{#1}

\bibitem[{Dao and Gu(2024)}]{dao2024ssm}
Tri Dao and Albert Gu. 2024.
\newblock \href {https://proceedings.mlr.press/v235/dao24a.html} {Transformers are {SSMs}: Generalized models and efficient algorithms through structured state space duality}.
\newblock In \emph{Proceedings of the 41st International Conference on Machine Learning}, volume 235 of \emph{Proceedings of Machine Learning Research}, pages 10041--10071. PMLR.

\bibitem[{{ggml-org}(2026)}]{llamacpp}
{ggml-org}. 2026.
\newblock \href {https://github.com/ggml-org/llama.cpp} {llama.cpp}.
\newblock Upstream inference runtime; experiments use a locally modified build.

\bibitem[{Gim et~al.(2024)Gim, Chen, Lee, Sarda, Khandelwal, and Zhong}]{gim2024promptcache}
In~Gim, Guojun Chen, Seung-seob Lee, Nikhil Sarda, Anurag Khandelwal, and Lin Zhong. 2024.
\newblock \href {https://proceedings.mlsys.org/paper_files/paper/2024/hash/a66caa1703fe34705a4368c3014c1966-Abstract-Conference.html} {Prompt cache: Modular attention reuse for low-latency inference}.
\newblock In \emph{Proceedings of Machine Learning and Systems}, volume~6.

\bibitem[{Hsieh et~al.(2024)Hsieh, Sun, Kriman, Acharya, Rekesh, Jia, Zhang, and Ginsburg}]{hsieh2024ruler}
Cheng-Ping Hsieh, Simeng Sun, Samuel Kriman, Shantanu Acharya, Dima Rekesh, Fei Jia, Yang Zhang, and Boris Ginsburg. 2024.
\newblock \href {https://arxiv.org/abs/2404.06654} {{RULER}: What's the real context size of your long-context language models?}
\newblock In \emph{First Conference on Language Modeling}.

\bibitem[{Hu et~al.(2025)Hu, Huang, Wang, Wang, Hu, Zhang, Feng, Chen, Shan, and Xie}]{hu2025epic}
Junhao Hu, Wenrui Huang, Weidong Wang, Haoyi Wang, Tiancheng Hu, Qin Zhang, Hao Feng, Xusheng Chen, Yizhou Shan, and Tao Xie. 2025.
\newblock \href {https://arxiv.org/abs/2410.15332} {{EPIC}: Efficient position-independent caching for serving large language models}.
\newblock \emph{arXiv preprint arXiv:2410.15332}.
\newblock Version 3.

\bibitem[{Jain and Wallace(2019)}]{jain2019attention}
Sarthak Jain and Byron~C. Wallace. 2019.
\newblock \href {https://doi.org/10.18653/v1/N19-1357} {Attention is not explanation}.
\newblock In \emph{Proceedings of the 2019 Conference of the North American Chapter of the Association for Computational Linguistics: Human Language Technologies, Volume 1 (Long and Short Papers)}, pages 3543--3556. Association for Computational Linguistics.

\bibitem[{Kwon et~al.(2023)Kwon, Li, Zhuang, Sheng, Zheng, Yu, Gonzalez, Zhang, and Stoica}]{kwon2023pagedattention}
Woosuk Kwon, Zhuohan Li, Siyuan Zhuang, Ying Sheng, Lianmin Zheng, Cody~Hao Yu, Joseph~E. Gonzalez, Hao Zhang, and Ion Stoica. 2023.
\newblock \href {https://arxiv.org/abs/2309.06180} {Efficient memory management for large language model serving with {PagedAttention}}.
\newblock In \emph{Proceedings of the 29th Symposium on Operating Systems Principles}. Association for Computing Machinery.

\bibitem[{Liu et~al.(2024)Liu, Lin, Hewitt, Paranjape, Bevilacqua, Petroni, and Liang}]{liu2024lostmiddle}
Nelson~F. Liu, Kevin Lin, John Hewitt, Ashwin Paranjape, Michele Bevilacqua, Fabio Petroni, and Percy Liang. 2024.
\newblock \href {https://doi.org/10.1162/tacl_a_00638} {Lost in the middle: How language models use long contexts}.
\newblock \emph{Transactions of the Association for Computational Linguistics}, 12:157--173.

\bibitem[{Liu et~al.(2026{\natexlab{a}})Liu, Wu, Liu, Hu, Li, Chen, and Chen}]{liu2026hypic}
Yifei Liu, Juntong Wu, Yang Liu, Junhao Hu, Minghao Li, Xiaoxu Chen, and Weihang Chen. 2026{\natexlab{a}}.
\newblock \href {https://arxiv.org/abs/2607.01299} {{HYPIC}: Accelerating hybrid-attention {LLM} serving with position-independent caching}.
\newblock \emph{arXiv preprint arXiv:2607.01299}.
\newblock Version 2.

\bibitem[{Liu et~al.(2026{\natexlab{b}})Liu, Qi, Wang, Wu, Chen, Jin, Shao, and Li}]{liu2026linearkv}
Yirui Liu, Ruoling Qi, Longwen Wang, Xuaner Wu, Jian Chen, Yuxin Jin, Jiawei Shao, and Xuelong Li. 2026{\natexlab{b}}.
\newblock \href {https://arxiv.org/abs/2608.11231} {{LinearKV}: One cached state suffices for position-independent caching in hybrid {LLMs}}.
\newblock \emph{arXiv preprint arXiv:2608.11231}.

\bibitem[{Maharana et~al.(2024)Maharana, Lee, Tulyakov, Bansal, Barbieri, and Fang}]{maharana2024locomo}
Adyasha Maharana, Dong-Ho Lee, Sergey Tulyakov, Mohit Bansal, Francesco Barbieri, and Yuwei Fang. 2024.
\newblock \href {https://doi.org/10.18653/v1/2024.acl-long.747} {Evaluating very long-term conversational memory of {LLM} agents}.
\newblock In \emph{Proceedings of the 62nd Annual Meeting of the Association for Computational Linguistics (Volume 1: Long Papers)}, pages 13851--13870. Association for Computational Linguistics.

\bibitem[{Packer et~al.(2023)Packer, Wooders, Lin, Fang, Patil, Stoica, and Gonzalez}]{packer2023memgpt}
Charles Packer, Sarah Wooders, Kevin Lin, Vivian Fang, Shishir~G. Patil, Ion Stoica, and Joseph~E. Gonzalez. 2023.
\newblock \href {https://arxiv.org/abs/2310.08560} {{MemGPT}: Towards {LLMs} as operating systems}.
\newblock \emph{arXiv preprint arXiv:2310.08560}.

\bibitem[{Park et~al.(2023)Park, O'Brien, Cai, Morris, Liang, and Bernstein}]{park2023generative}
Joon~Sung Park, Joseph~C. O'Brien, Carrie~J. Cai, Meredith~Ringel Morris, Percy Liang, and Michael~S. Bernstein. 2023.
\newblock \href {https://doi.org/10.1145/3586183.3606763} {Generative agents: Interactive simulacra of human behavior}.
\newblock In \emph{Proceedings of the 36th Annual ACM Symposium on User Interface Software and Technology}. Association for Computing Machinery.

\bibitem[{{Qwen Team}(2026)}]{qwen2026model}
{Qwen Team}. 2026.
\newblock \href {https://huggingface.co/Qwen/Qwen3.6-27B} {{Qwen3.6-27B}: Model card}.
\newblock Accessed 16 September 2026. Base-family reference; experiments use a derivative quantized checkpoint.

\bibitem[{Su et~al.(2021)Su, Lu, Pan, Murtadha, Wen, and Liu}]{su2021roformer}
Jianlin Su, Yu~Lu, Shengfeng Pan, Ahmed Murtadha, Bo~Wen, and Yunfeng Liu. 2021.
\newblock \href {https://arxiv.org/abs/2104.09864} {{RoFormer}: Enhanced transformer with rotary position embedding}.
\newblock \emph{arXiv preprint arXiv:2104.09864}.

\bibitem[{Wang et~al.(2023)Wang, Dong, Cheng, Liu, Yan, Gao, and Wei}]{wang2023longmem}
Weizhi Wang, Li~Dong, Hao Cheng, Xiaodong Liu, Xifeng Yan, Jianfeng Gao, and Furu Wei. 2023.
\newblock \href {https://proceedings.neurips.cc/paper_files/paper/2023/hash/ebd82705f44793b6f9ade5a669d0f0bf-Abstract-Conference.html} {Augmenting language models with long-term memory}.
\newblock In \emph{Advances in Neural Information Processing Systems}, volume~36.

\bibitem[{Wiegreffe and Pinter(2019)}]{wiegreffe2019attention}
Sarah Wiegreffe and Yuval Pinter. 2019.
\newblock \href {https://doi.org/10.18653/v1/D19-1002} {Attention is not not explanation}.
\newblock In \emph{Proceedings of the 2019 Conference on Empirical Methods in Natural Language Processing and the 9th International Joint Conference on Natural Language Processing (EMNLP-IJCNLP)}, pages 11--20. Association for Computational Linguistics.

\bibitem[{Wu et~al.(2025)Wu, Wang, Yu, Zhang, Chang, and Yu}]{wu2025longmemeval}
Di~Wu, Hongwei Wang, Wenhao Yu, Yuwei Zhang, Kai-Wei Chang, and Dong Yu. 2025.
\newblock \href {https://arxiv.org/abs/2410.10813} {{LongMemEval}: Benchmarking chat assistants on long-term interactive memory}.
\newblock In \emph{The Thirteenth International Conference on Learning Representations}.

\bibitem[{Xiao et~al.(2024)Xiao, Tian, Chen, Han, and Lewis}]{xiao2024streamingllm}
Guangxuan Xiao, Yuandong Tian, Beidi Chen, Song Han, and Mike Lewis. 2024.
\newblock \href {https://openreview.net/forum?id=NG7sS51zVF} {Efficient streaming language models with attention sinks}.
\newblock In \emph{The Twelfth International Conference on Learning Representations}.

\bibitem[{Yang et~al.(2025)Yang, Kautz, and Hatamizadeh}]{yang2025gated}
Songlin Yang, Jan Kautz, and Ali Hatamizadeh. 2025.
\newblock \href {https://arxiv.org/abs/2412.06464} {Gated delta networks: Improving {Mamba2} with delta rule}.
\newblock In \emph{The Thirteenth International Conference on Learning Representations}.

\bibitem[{Yao et~al.(2025)Yao, Li, Liu, Ray, Cheng, Zhang, Du, Lu, and Jiang}]{yao2025cacheblend}
Jiayi Yao, Hanchen Li, Yuhan Liu, Siddhant Ray, Yihua Cheng, Qizheng Zhang, Kuntai Du, Shan Lu, and Junchen Jiang. 2025.
\newblock \href {https://arxiv.org/abs/2405.16444} {{CacheBlend}: Fast large language model serving for {RAG} with cached knowledge fusion}.
\newblock \emph{arXiv preprint arXiv:2405.16444}.
\newblock Version 3.

\bibitem[{Zheng et~al.(2023)Zheng, Chiang, Sheng, Zhuang, Wu, Zhuang, Lin, Li, Li, Xing, Zhang, Gonzalez, and Stoica}]{zheng2023judge}
Lianmin Zheng, Wei-Lin Chiang, Ying Sheng, Siyuan Zhuang, Zhanghao Wu, Yonghao Zhuang, Zi~Lin, Zhuohan Li, Dacheng Li, Eric~P. Xing, Hao Zhang, Joseph~E. Gonzalez, and Ion Stoica. 2023.
\newblock \href {https://proceedings.neurips.cc/paper_files/paper/2023/hash/91f18a1287b398d378ef22505bf41832-Abstract-Datasets_and_Benchmarks.html} {Judging {LLM}-as-a-judge with {MT-Bench} and {Chatbot Arena}}.
\newblock In \emph{Advances in Neural Information Processing Systems}, volume~36.

\bibitem[{Zheng et~al.(2024)Zheng, Yin, Xie, Sun, Huang, Yu, Cao, Kozyrakis, Stoica, Gonzalez, Barrett, and Sheng}]{zheng2024sglang}
Lianmin Zheng, Liangsheng Yin, Zhiqiang Xie, Chuyue Sun, Jeff Huang, Cody~Hao Yu, Shiyi Cao, Christos Kozyrakis, Ion Stoica, Joseph~E. Gonzalez, Clark Barrett, and Ying Sheng. 2024.
\newblock \href {https://proceedings.neurips.cc/paper_files/paper/2024/hash/724be4472168f31ba1c9ac630f15dec8-Abstract-Conference.html} {{SGLang}: Efficient execution of structured language model programs}.
\newblock In \emph{Advances in Neural Information Processing Systems}, volume~37.

\bibitem[{Zhong et~al.(2024)Zhong, Guo, Gao, Ye, and Wang}]{zhong2024memorybank}
Wanjun Zhong, Lianghong Guo, Qiqi Gao, He~Ye, and Yanlin Wang. 2024.
\newblock \href {https://doi.org/10.1609/aaai.v38i17.29946} {{MemoryBank}: Enhancing large language models with long-term memory}.
\newblock In \emph{Proceedings of the AAAI Conference on Artificial Intelligence}, volume~38, pages 19724--19731.

\end{thebibliography}

\appendix
\section{Decisive Case Excerpts}
\label{app:examples}
The following excerpts are selected diagnostic examples, not a random sample or a complete success/failure count. English passages are translations unless a schema is shown. Chinese originals are retained where a participant or temporal binding is decisive.

\subsection{Independent composition: event time}
\quotebox{Recorded memory}{\cn{今日卯时。天亮后，王婶在自家门口附近发现了尹旭。}\\
Today at dawn. After daybreak, Aunt Wang discovered Yin Xu near her own doorway.}
\quotebox{Independent-composition response plan}{\cn{简要解释他昨夜在王婶家门口被发现，受了伤且暂时失忆，强调他现在是安全的，无需害怕。}\\
Briefly explain that he was found at Aunt Wang's doorway last night, injured and temporarily amnesiac; emphasize that he is safe now and need not be afraid.}
\quotebox{Full-refill dialogue}{\cn{今晨……王婶在自家门口附近发现了你。}\\
This morning ... Aunt Wang found you near her own doorway.}
The error is the planner's explicit placement of the discovery last night. We do not rely on an ellipsis in the independent final dialogue to classify the result. Another independent plan suggests that the player woke at the doorway; the available details allow that interpretation, so it is not counted here as an unambiguous error.

\subsection{Eight updates: present danger, retained kindness}
A relative who once rescued Lin Qingyao later attacks the household. The current cognition marks him dangerous; the older rescue is retained in a merged memory. The maintained dialogue distinguishes both:
\quotebox{Maintained dialogue}{\cn{过去的温情是真的，现在的危险也是真的。}\\
The warmth in the past was real; the danger now is real too.}
This example illustrates revision without treating the earlier experience as false. A later target binding uses the existing relative ID. It does not validate the unsupported cultivation rank added by the final dialogue.

\subsection{Current quantity versus completed expenditure}
The revised memory records three herbs collected, one given to Granny He, and two remaining. The current quantity is two. The request asks for the remainder, excluding a separate sleeping-aid item.
\quotebox{Tail-and-relocate compilation explanation}{\cn{根据记忆，我共有两株暖脉草，昨日已送一株给何婆婆，故剩余一株。}\\
According to memory, I have two warm-pulse herbs in total; I gave one to Granny He yesterday, so one remains.}
The true-tail reconstruction correctly uses two as the post-transfer quantity. The underlying mixed replay's concrete binding is:
\begin{quote}\small\ttfamily
\{"kind":"entity",\\
"entity\_type":"item",\\
"id":"item\_warm\_pulse\_grass",\\
"quantity":2,"unit":""\}
\end{quote}
The stale ``not currently held'' phrase described in the main text is shared by all matched placement fixtures. It should be retained in the archived fixture rather than silently cleaned while reporting these outputs.

\subsection{Long-horizon affect: recovering the event actor}
\label{app:prompt-recovery}
The player asks who collected the scattered papers and who later held them. The frozen cognition states that Yin Xu picked up the pages, cleaned the muddy ones, and that the notebook had been taken back. The changed prompt contains no replacement facts.
\quotebox{Original-task control, excerpt}{\cn{记得是我捡回散页}\\
I remember that I picked up the scattered pages.}
\quotebox{Understanding-first task, excerpt}{\cn{那日是他细心捡回并替我拍净}\\
That day, he carefully picked them up and brushed them clean for me.}
Here ``I'' is Lin Qingyao and ``he'' refers to Yin Xu. The control also confuses the notebook episode with a red cord; the revised output does not. The affective explanation remains subjective and relatively long, but the objective participant binding is restored. These excerpts concern the affect module, not a rerun of the complete downstream conversation.

\section{Maintenance Trace and Audit Targets}
\begin{table*}[t]
\centering\small
\begin{tabularx}{\textwidth}{@{}cp{0.39\textwidth}Y@{}}
\toprule
Update & Revised cognition & Memory consolidation \\
\midrule
1 & Charm: seller's claim versus brief observed relief & Merge purchase and trial; remove stall detail. \\
2 & Herb location discovered; holding count rises to three & Merge identification and collection; remove recognition process. \\
3 & Charm worsens the curse; seller's reliability revised & Retain claim, observation, and contradiction as distinct sources. \\
4 & Herb slope has dusk poison fog; ledger identifies the herb & Merge resource, risk, and ledger; remove route detail. \\
5 & Cord delivery switches to Granny He; Aunt Wang only relays & Merge custody and revised recipient; remove original meeting place. \\
6 & Rescuer becomes an attacker; family home is unsafe & Retain old kindness and new violence; remove sleeve detail. \\
7 & Old shrine offers shelter; one herb given, two remain & Merge gathering, accounting, expenditure, and shelter. \\
8 & Charm retained as evidence; other herb only aids sleep & Compress repeated tests; preserve contrasting item effects. \\
\bottomrule
\end{tabularx}
\caption{Eight scripted maintenance steps. The initially shuffled 25 records become 24 live records. Each step updates one or two cognition entries together with related memories; unchanged older experiences remain in their original KV.}
\end{table*}

\paragraph{Suggested audit record.}
For later expansion, record the source fact, maintenance version, relevant request, module output, and adjudication rationale together. Keep at least four categories separate: supported, defensible inference, explicit conflict, and insufficient evidence. An output may be mechanically correct but awkwardly expressed, or fluent while mechanically wrong. Count these separately. This is an evaluation plan, not a completed judge-based benchmark.

\section{Attention Definitions and Caveats}
The first eight tokens of each memory group define the boundary-window diagnostic. In the 34-group experiment these windows occupy 2.05\% of memory tokens; they do not encompass every entire header. Summary/detail text values constitute the body category. IDs, field names, delimiters, and time metadata are classified separately, with small ambiguity where a token crosses a field boundary.

Reference-softmax tracing uses the queries actually computed in each state, not one identical numerical query applied to different keys. Therefore recurrence can affect the measured distribution. Fixed continuations avoid differences in generated text but do not turn attention mass into a causal attribution. Head aggregation can conceal specialized behavior, and the reference computation can differ in rounding from fused CUDA kernels.

The 10K heatmaps use within-record attention normalization per layer, with an explicitly marked color range. Absolute destination shares and token-level distribution distances are reported separately. The true-tail and key-relocation behavior runs do not have corresponding captured attention maps in the existing artifacts. No such missing maps are inferred from the other cohorts.

\section{Evidence Provenance}
\label{app:provenance}
The study draws on local experimental reports and replay artifacts dated 13--16 September 2026. The cohort identifiers below distinguish experimental histories and diagnostic runs; they are not public download locations. The accompanying numerical snapshots document the plotted measurements and calibrations. Some early binary state dumps were deleted to reclaim storage, while derived attention arrays, figures, logs, and summaries were retained. These snapshots do not reconstruct the full runtime experiments or unavailable state dumps.

\begin{description}
\item[Independent replay:] \code{20260913\_060631}; two turns, shared first request, own-history second turns.
\item[Independent attention:] \code{memory\_attention/20260913}; frozen first-turn suffix and reference continuation.
\item[Five-update attention:] \code{current\_20260913}; matched final 38-group memory text, four probes.
\item[Mixed-record replay:] \code{20260916\_014747}; eight updates, nine complete dialogue turns. The earlier harness-failure attempt is excluded.
\item[Three-way attention:] \code{20260916\_threeway}; frozen dialogue, herb, and gate requests.
\item[True-tail ablation:] \code{20260916\_041622}; matched chat endpoint, three reconstructions. An earlier endpoint-mismatched trial is excluded.
\item[Rotate-back ablation:] \code{20260916\_043418}; three reconstructions, keys rotated after tail computation.
\end{description}

\paragraph{Long-horizon prompt probes:} \code{attention\_ab\_20260916}; frozen task-only pairs on the restored long-horizon root after a one-hour time update. The accompanying snapshot records the controlled comparison and quoted excerpts.

\paragraph{Accompanying data and availability.}
The source package includes three numerical snapshots in \code{anc/}: \code{plot\_data.json} for attention diagnostics, \code{performance\_data.json} for timing measurements and calibrated estimates, and \code{prompt\_recovery.json} for the frozen-state prompt comparison. Artifact references identify the original records without exposing local filesystem paths. These are supporting data, not a release of the game implementation, full prompting strategy, or binary model states. Multi-character evaluation and further controlled experiments remain future work.

\end{document}